\documentclass[11pt,a4paper]{article}
\usepackage[hyperref]{naaclhlt2018}
\usepackage{times}
\usepackage{latexsym}
\usepackage{pgfplots}
\pgfplotsset{compat=1.8}
\usepackage{multirow}

\usepackage{url}

\aclfinalcopy

\title{Multi-Source Syntactic Neural Machine Translation}

\author{Anna Currey \\
        University of Edinburgh \\
        {\tt a.currey@sms.ed.ac.uk}
  \\\And
  Kenneth Heafield \\
  University of Edinburgh \\
  {\tt kheafiel@inf.ed.ac.uk}}

\date{}

\begin{document}
\maketitle
\begin{abstract}
We introduce a novel multi-source technique for incorporating source syntax into neural machine translation using linearized parses. This is achieved by employing separate encoders for the sequential and parsed versions of the same source sentence; the resulting representations are then combined using a hierarchical attention mechanism. The proposed model improves over both seq2seq and parsed baselines by over 1 BLEU on the WMT17 English$\rightarrow$German task. Further analysis shows that our multi-source syntactic model is able to translate successfully without any parsed input, unlike standard parsed methods. In addition, performance does not deteriorate as much on long sentences as for the baselines.
\end{abstract}

\section{Introduction}
\label{sect:intro}
Neural machine translation (NMT) typically makes use of a recurrent neural network (RNN) -based encoder and decoder, along with an attention mechanism~\cite{bahdanau2014neural,cho2014learning,kalchbrenner2013recurrent,sutskever2014sequence}. However, it has been shown that RNNs require some supervision to learn syntax~\cite{bentivogli2016neural,linzen2016assessing,shi2016does}. Therefore, explicitly incorporating syntactic information into NMT has the potential to improve performance. This is particularly true for source syntax, which can improve the model's representation of the source language.

Recently, there have been a number of proposals for using linearized representations of parses within standard NMT~\cite{aharoni2017towards,li2017modeling,nadejde2017syntax}. Linearized parses are advantageous because they can inject syntactic information into the models without significant changes to the architecture. However, using linearized parses in a sequence-to-sequence (seq2seq) framework creates some challenges, particularly when using source parses. First, the parsed sequences are significantly longer than standard sentences, since they contain node labels as well as words. Second, these systems often fail when the source sentence is not parsed. This can be a problem for inference, since the external parser may fail on an input sentence at test time. We propose a method for incorporating linearized source parses into NMT that addresses these challenges by taking both the sequential source sentence and its linearized parse simultaneously as input in a multi-source framework. Thus, the model is able to use the syntactic information encoded in the parse while falling back to the sequential sentence when necessary. Our proposed model improves over both standard and parsed NMT baselines.

\section{Related Work}

\label{sect:related}
\subsection{Seq2seq Neural Parsing}
\label{sect:related-parse}
Using linearized parse trees within sequential frameworks was first done in the context of neural parsing. \newcite{vinyals2015grammar} parsed using an attentional seq2seq model; they used linearized, unlexicalized parse trees on the target side and sentences on the source side. In addition, as in this work, they used an external parser to create synthetic parsed training data, resulting in improved parsing performance. 
\newcite{choe2016parsing} adopted a similar strategy, using linearized parses in an RNN language modeling framework.

\subsection{NMT with Source Syntax}
\label{sect:related-sourcesyn}
Among the first proposals for using source syntax in NMT was that of \newcite{luong2015multi}, who introduced a multi-task system in which the source data was parsed and translated using a shared encoder and two decoders. 
More radical changes to the standard NMT paradigm have also been proposed. \newcite{eriguchi2016tree} introduced tree-to-sequence NMT; this model took parse trees as input using a tree-LSTM~\cite{tai2015improved} encoder. \newcite{bastings2017graph} used a graph convolutional encoder in order to take labeled dependency parses of the source sentences into account. \newcite{hashimoto2017neural} added a latent graph parser to the encoder, allowing it to learn soft dependency parses while simultaneously learning to translate.

\subsection{Linearized Parse Trees in NMT}
\label{sect:related-linparse}
The idea of incorporating linearized parses into seq2seq has been adapted to NMT as a means of injecting syntax. \newcite{aharoni2017towards} first did this by parsing the target side of the training data and training the system to generate parsed translations of the source input; this is the inverse of our parse2seq baseline. Similarly, \newcite{nadejde2017syntax} interleaved CCG supertags with words on the target side, finding that this improved translation despite requiring longer sequences.

Most similar to our multi-source model is the parallel RNN model proposed by \newcite{li2017modeling}. Like multi-source, the parallel RNN used two encoders, one for words and the other for syntax. However, they combined these representations at the word level, whereas we combine them on the sentence level. Their mixed RNN model is also similar to our parse2seq baseline, although the mixed RNN decoder attended only to words. As the mixed RNN model outperformed the parallel RNN model, we do not attempt to compare our model to parallel RNN. These models are similar to ours in that they incorporate linearized parses into NMT; here, we utilize a multi-source framework.

\subsection{Multi-Source NMT}
\label{sect:related-ms}
Multi-source methods in neural machine translation were first introduced by~\newcite{zoph2016multi} for multilingual translation. They used one encoder per source language, and combined the resulting sentence representations before feeding them into the decoder. \newcite{firat2016multi} expanded on this by creating a multilingual NMT system with multiple encoders and decoders. \newcite{libovicky2017attention} applied multi-source NMT to multimodal translation and automatic post-editing and explored different strategies for combining attention over the two sources. In this paper, we apply the multi-source framework to a novel task, syntactic neural machine translation.

\section{NMT with Linearized Source Parses}
\label{sect:model}
We propose a multi-source method for incorporating source syntax into NMT. This method makes use of linearized source parses; we describe these parses in section~\ref{sect:model-linparse}. Throughout this paper, we refer to standard sentences that do not contain any explicit syntactic information as \textit{sequential}; see Table~\ref{tab:ex_sentences} for an example.

\subsection{Linearized Source Parses}
\label{sect:model-linparse}
We use an off-the-shelf parser, in this case Stanford CoreNLP~\cite{manning2014stanford}, to create binary constituency parses. These parses are linearized as shown in Table~\ref{tab:ex_sentences}.  
We tokenize the opening parentheses with the node label (so each node label begins with a parenthesis) but keep the closing parentheses separate from the words they follow.  For our task, the parser failed on one training sentence of 5.9 million, which we discarded, and succeeded on all test sentences. It took roughly 16 hours to parse the 5.9 million training sentences.

Following \newcite{sennrich2015neural}, our networks operate at the subword level using byte pair encoding (BPE) with a shared vocabulary on the source and target sides.  However, the parser operates at the word level.  Therefore, we parse then break into subwords, so a leaf may have multiple tokens without internal structure.  

The proposed method is tested using both \textit{lexicalized} and \textit{unlexicalized} parses. In \textit{unlexicalized} parses, we remove the words, keeping only the node labels and the parentheses. In \textit{lexicalized} parses, the words are included. Table~\ref{tab:ex_sentences} shows an example of the three source sentence formats: sequential, lexicalized parse, and unlexicalized parse. Note that the lexicalized parse is significantly longer than the other versions.

\begin{table*}
\small\centering
\begin{tabular}{|l|l|} \hline
\bf & \bf Example Sentence \\\hline
sequential & history is a great teacher .\\
lexicalized parse & (ROOT (S (NP (NN history ) ) (VP (VBZ is ) (NP (DT a ) (JJ great ) (NN teacher ) ) ) (. . ) ) )\\
unlexicalized parse & (ROOT (S (NP (NN ) ) (VP (VBZ ) (NP (DT ) (JJ ) (NN ) ) ) (. . ) ) )\\\hline
target sentence & die Geschichte ist ein gro{\ss}artiger Lehrmeister .\\
\hline
\end{tabular}
\caption{\label{tab:ex_sentences} Example source training sentence with sequential, lexicalized parse, and unlexicalized parse versions. We include the corresponding target sentence for reference.}
\end{table*}

\subsection{Multi-Source}
\label{sect:model-multisource}
We propose a multi-source framework for injecting linearized source parses into NMT. This model consists of two identical RNN encoders with no shared parameters, as well as a standard RNN decoder. For each target sentence, two versions of the source sentence are used: the sequential (standard) version and the linearized parse (lexicalized or unlexicalized). Each of these is encoded simultaneously using the encoders; the encodings are then combined and used as input to the decoder. We combine the source encodings using the hierarchical attention combination proposed by \newcite{libovicky2017attention}. This consists of a separate attention mechanism for each encoder; these are then combined using an additional attention mechanism over the two separate context vectors. This multi-source method is thus able to combine the advantages of both standard RNN-based encodings and syntactic encodings.

\section{Experimental Setup}
\label{sect:exp}
\subsection{Data}
\label{sect:exp-data}
We base our experiments on the WMT17 \cite{bojar2017findings} English (EN) $\rightarrow$ German (DE) news translation task. All 5.9 million parallel training sentences are used, but no monolingual data. Validation is done on newstest2015, while newstest2016 and newstest2017 are used for testing.

We train a shared BPE vocabulary with 60k merge operations on the parallel training data. For the parsed data, we break words into subwords after applying the Stanford parser. We tokenize and truecase the data using the Moses tokenizer and truecaser~\cite{koehn2007moses}.

\subsection{Implementation}
\label{sect:exp-imp}
The models are implemented in Neural Monkey~\cite{helcl2017neural}. They are trained using Adam~\cite{kingma2014adam} and have minibatch size 40, RNN size 512, and dropout probability 0.2~\cite{gal2016theoretically}. We train to convergence on the validation set, using BLEU~\cite{papineni2002bleu} as the metric.

For sequential inputs and outputs, the maximum sentence length is 50 subwords. For parsed inputs, we increase maximum sentence length to 150 subwords to account for the increased length due to the parsing labels; we still use a maximum output length of 50 subwords for these systems.

\subsection{Baselines}
\label{sect:exp-base}
\subsubsection*{Seq2seq}
\label{sect:exp-base-s2s}
The proposed models are compared against two baselines. The first, referred to here as \textit{seq2seq}, is the standard RNN-based neural machine translation system with attention~\cite{bahdanau2014neural}. This baseline does not use the parsed data.

\subsubsection*{Parse2seq}
\label{sect:exp-base-p2s}
The second baseline we consider is a slight modification of the mixed RNN model proposed by \newcite{li2017modeling}. This uses an identical architecture to the seq2seq baseline (except for a longer maximum sentence length in the encoder). Instead of using sequential data on the source side, the linearized parses are used. We allow the system to attend equally to words and node labels on the source side, rather than restricting the attention to words. We refer to this baseline as \textit{parse2seq}.

\section{Results}
\label{sect:res}

Table~\ref{tab:res_all} shows the performance on EN$\rightarrow$DE translation for each of the proposed systems and the baselines, as approximated by BLEU score. 

\begin{table}
\small\centering
\begin{tabular}{|l|l|cc|} \hline
  &\bfseries System
  & \textbf{2016} & \textbf{2017}\\
  \hline
\multirow{2}{*}{baseline}&seq2seq &     25.0 & 20.8 \\
&parse2seq &    25.4 & 20.9 \\
\hline
\multirow{2}{*}{proposed}&multi-source lex &    \bf 26.5 & \bf 21.9 \\
&multi-source unlex &    26.4 & 21.7 \\
\hline
\end{tabular}
\caption{\label{tab:res_all} BLEU scores on newstest2016 and newstest2017 datasets for the baselines, unlexicalized (unlex), and lexicalized (lex) systems.}
\end{table}

The multi-source systems improve strongly over both baselines, with improvements of up to 1.5 BLEU over the seq2seq baseline and up to 1.1 BLEU over the parse2seq baseline. In addition, the lexicalized multi-source systems yields slightly higher BLEU scores than the unlexicalized multi-source systems; this is surprising because the lexicalized systems have significantly longer sequences than the unlexicalized ones. Finally, it is interesting to compare the seq2seq and parse2seq baselines. Parse2seq outperforms seq2seq by only a small amount compared to multi-source; thus, while adding syntax to NMT can be helpful, some ways of doing so are more effective than others.

\section{Analysis}
\label{sect:qual}

\subsection{Inference Without Parsed Sentences}
\label{sect:qual-parse}
The parse2seq and multi-source systems require parsed source data at inference time. However, the parser may fail on an input sentence. Therefore, we examine how well these systems do when given only unparsed source sentences at test time.

Table~\ref{tab:res_multisource} displays the results of these experiments. For the parse2seq baseline, we use only sequential (\textit{seq}) data as input. For the lexicalized and unlexicalized multi-source systems, two options are considered: \textit{seq + seq} uses identical sequential data as input to both encoders, while \textit{seq + null} uses null input for the parsed encoder, where every source sentence is ``( )''.

The parse2seq system fails when given only sequential source data. On the other hand, both multi-source systems perform reasonably well without parsed data, although the BLEU scores are worse than multi-source with parsed data.

\begin{table}
\small\centering
\begin{tabular}{|l|l|cc|} \hline
\bfseries System & \bf Source Data  & \textbf{2016} & \textbf{2017}\\\hline
parse2seq & seq & 0.6 & 0.5\\\hline
multi-source lex & seq + seq & 23.6 & 20.0 \\
 & seq + null & 23.1 & 19.3 \\\hline
multi-source unlex & seq + seq & \bf 23.7 & 19.9 \\
 & seq + null & 23.6 & \bf 20.9 \\
\hline
\end{tabular}
\caption{\label{tab:res_multisource} BLEU scores on newstest2016 and newstest2017 when no parsed data is used during inference.}
\end{table}

\subsection{BLEU by Sentence Length}
\label{sect:qual-len}
For models that use source-side linearized parses (multi-source and parse2seq), the source sequences are significantly longer than for the seq2seq baseline. Since NMT already performs relatively poorly on long sentences~\cite{bahdanau2014neural}, adding linearized source parses may exacerbate this issue. To detect whether this occurs, we calculate BLEU by sentence length.

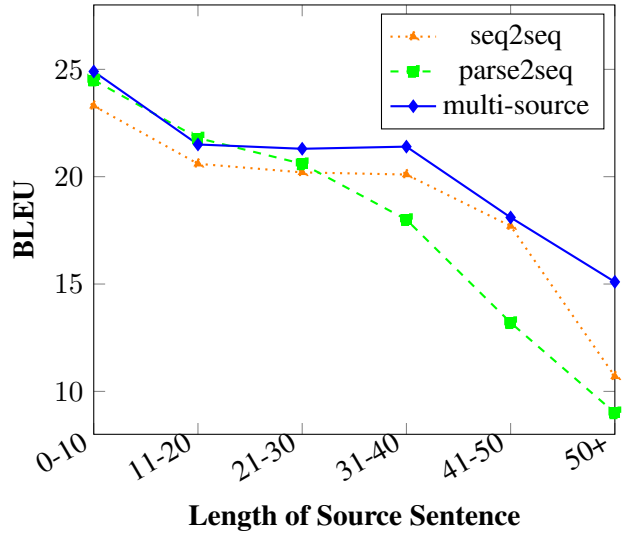
\begin{figure}
\begin{tikzpicture}
\begin{axis}[
  ymin=0,
  ylabel=\textbf{BLEU},
  xlabel=\textbf{Length of Source Sentence},
  ymin=8,ymax=28,
  xticklabel style = {rotate=30,anchor=east},
  enlargelimits = false,
  xticklabels from table={2017bylength.tex}{Length},xtick=data
]
\addplot[orange,thick,dotted,mark=triangle*] table [y=seq2seq,x=X]{2017bylength.tex};
\addlegendentry{seq2seq}
\addplot[green,thick,dashed,mark=square*] table [y=parse2seq,x=X]{2017bylength.tex};
\addlegendentry{parse2seq}
\addplot[blue,thick,mark=diamond*] table [y=multi-source,x=X]{2017bylength.tex};
\addlegendentry{multi-source}
\end{axis}
\end{tikzpicture}
\caption{\label{tab:res_len} BLEU by sentence length on newstest2017 for baselines and lexicalized multi-source.}
\end{figure}

We bucket the sentences in newstest2017 by source sentence length. We then compute BLEU scores for each bucket for the seq2seq and parse2seq baselines and the lexicalized multi-source system. The results are in Figure~\ref{tab:res_len}. 

In line with previous work on NMT on long sentences~\cite{bahdanau2014neural,li2017modeling}, we see a significant deterioration in BLEU for longer sentences for all systems. In particular, although the parse2seq model outperformed the seq2seq model overall, it does worse than seq2seq for sentences containing more than 30 words. This indicates that parse2seq performance does indeed suffer due to its long sequences. On the other hand, the multi-source system outperforms the seq2seq baseline for all sentence lengths and does particularly well for sentences with over 50 words. This may be because the multi-source system has both sequential and parsed input, so it can rely more on sequential input for very long sentences.

\section{Conclusion}
\label{sect:concl}
In this paper, we presented a multi-source method for effectively incorporating linearized parses of the source data into neural machine translation. This method, in which the parsed and sequential versions of the sentence were both taken as input during training and inference, resulted in gains of up to 1.5 BLEU on EN$\rightarrow$DE translation. In addition, unlike parse2seq, the multi-source model translated reasonably well even when the source sentence was not parsed.

In the future, we will explore adding back-translated~\cite{sennrich2015improving} or copied~\cite{currey2017copied} target data to our multi-source system. The multi-source model does not require all training data to be parsed; thus, monolingual data can be used even if the parser is unreliable for the synthetic or copied source sentences.

\section*{Acknowledgments}
This work was funded by the Amazon Academic Research Awards program.

\bibliography{phdrefs}
\bibliographystyle{acl_natbib}

\end{document}